# Human Image Generation: A Comprehensive Survey

ZHEN JIA, ZHANG ZHANG*, LIANG WANG*, and TIENIU TAN, New Laboratory of Pattern Recognition (NLPR), State Key Laboratory of Multimodal Artificial Intelligence Systems (MAIS), Institute of Automation, Chinese Academy of Sciences, China

Image and video synthesis has become a blooming topic in computer vision and machine learning communities along with the developments of deep generative models, due to its great academic and application value. Many researchers have been devoted to synthesizing high-fidelity human images as one of the most commonly seen object categories in daily lives, where a large number of studies are performed based on various models, task settings and applications. Thus, it is necessary to give a comprehensive overview on these variant methods on human image generation. In this paper, we divide human image generation techniques into three paradigms, i.e., data-driven methods, knowledge-guided methods and hybrid methods. For each paradigm, the most representative models and the corresponding variants are presented, where the advantages and characteristics of different methods are summarized in terms of model architectures. Besides, the main public human image datasets and evaluation metrics in the literature are summarized. Furthermore, due to the wide application potentials, the typical downstream usages of synthesized human images are covered. Finally, the challenges and potential opportunities of human image generation are discussed to shed light on future research.

## 1 INTRODUCTION

Over the past decade, significant progress has been achieved in image synthesis filed, driven by the prosperousness of deep generative models, e.g., generative adversarial networks (GANs) [18, 49, 77, 227] and variational autoencoders (VAEs) [79, 156], which attracts researchers to develop various visual generation tasks. Among these tasks, human image generation, also termed human image synthesis and person image generation, is becoming a flourishing research topic, due to its academic research values and great application potentials. Specially, human image generation models should simulate complicated imaging variations inherent in human bodies, encompassing articulated and non-rigid deformations, diverse clothing attributes, and the complex interplay of background, illumination, and viewpoints across different camera perspectives. Meanwhile, the vast potential applications such as clothing virtual try-on (VITON), animated films and games,

*Corresponding authors of this paper.

Authors' address: Zhen Jia, zhen.jia@nlpr.ia.ac.cn; Zhang Zhang, zzhang@nlpr.ia.ac.cn; Liang Wang, wangliang@nlpr.ia.ac.cn; Tieniu Tan, tnt@nlpr.ia.ac.cn, New Laboratory of Pattern Recognition (NLPR), State Key Laboratory of Multimodal Artificial Intelligence Systems (MAIS), Institute of Automation, Chinese Academy of Sciences, Beijing, China, 100190.

and the burgeoning demand for virtual human avatar services in the metaverse, also inspire researchers developing various human image generation algorithms. Human image generation has become an open and booming research frontier in computer vision (CV) and computer graphics (CG) communities.

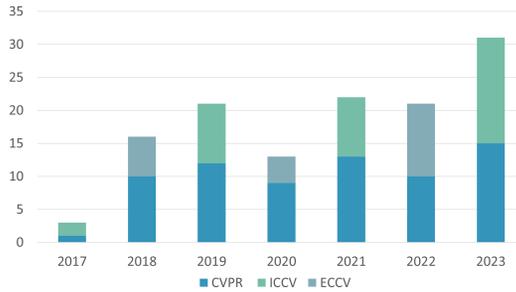

Fig. 1. The number of papers of human image generation on the top computer vision conferences (CVPR, ICCV and ECCV) since 2017.

Figure 1 illustrates the trend in the number of papers on human image generation published in top computer vision conferences (CVPR, ICCV and ECCV) in recent years, which shows the growing interest and emphasis on human image synthesis following the widespread adoption of deep neural networks [49, 62, 79, 114, 166]. Simultaneously, the development of numerous large-scale human image datasets [28, 56, 90, 104, 191, 221], which serve as invaluable resources by providing extensive training data and establishing standard evaluation benchmarks, has also significantly contributed to this surge in research activity.

In this survey, more than 200 papers focusing on human image generation are reviewed, primarily sourced from from well-known academic conferences and journals, such as CVPR, ICCV, ECCV, NeurIPS, TPAMI, TIP, 3DV and TOG over the past decade. We aim to furnish a comprehensive and insightful review of human image generation, delving into diverse models, task settings, evaluation benchmarks and applications, with a particular emphasis on synthesizing complete human bodies. Specifically, we review main literatures on human image generation in the past decade and divide them into three generation paradigms in terms of various methods, i.e., **data-driven methods, knowledge-guided methods** and **hybrid methods**. This classification facilitates a structured exploration of various approaches, enabling a deeper understanding of the evolving landscape of human image generation research.

The data-driven methods, such as [15, 42, 56, 86, 108, 141], represent a bottom-up paradigm leveraging certain deep neural networks, e.g., GAN [49], VAE [79], U-Net [140], Transformer [166], NeRF [114] and diffusion models [62], to generate novel human images from learned training data. In the exploration of data-driven methods, we propose two method taxonomies based on fundamental models and task settings. Subsequently, we introduce and analyze an overall pipeline of data-driven methods, comprising three key components: feature encoding, image generation, and loss functions. Within each component, we delve into the common strategies and frameworks employed in detail.

Knowledge-guided methods, as seen in works like [116, 159, 191], denote a top-down paradigm for generating human images by leveraging rich human prior knowledge, including 3D models of human bodies [6, 7, 105, 124, 126], human kinematics [53], and appearance priors [70] in imaging process. In the exploration of knowledge-guided methods, we begin by introducing commonly used fundamental models. Subsequently, we categorize and introduce methods through two primary pipelines: the pixel warping pipeline and the virtual rendering pipeline.

Hybrid methods, as evidenced by numerous recent studies such as [86, 145, 193, 204], capitalize on the complementary advantages of both data-driven and knowledge-guided approaches. In these methods, 3D geometry and shape models serve as essential inputs to data-driven model pipelines, facilitating the learning of complex appearance variations in the training data. Figure 2 illustrates the sketch diagrams of the three paradigms.

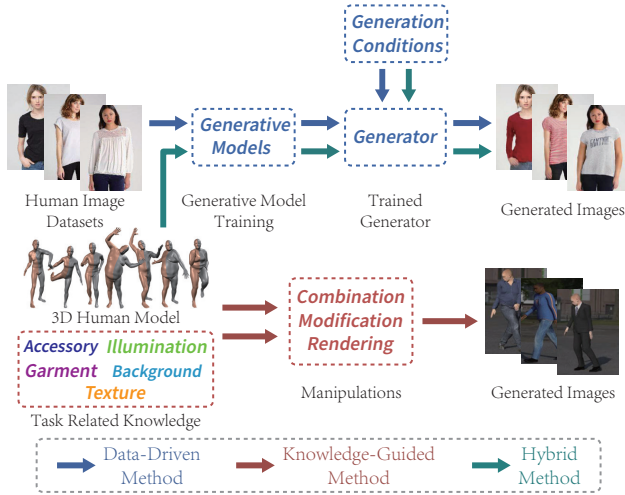

Fig. 2. The sketch diagrams of three paradigms in human image generation, i.e., data-driven methods, knowledge-guided methods, and hybrid methods.

Recently, several surveys have delved into specific subtopics in human image generation. For instance, Liu et al. [99] offer an overview of algorithms and applications centered around GAN models, covering tasks such as human pose transfer and human body rendering. Similarly, Cheng et al. [26] present a comprehensive survey on the technical progress in fashion applications, highlighting fashion synthesis as one of the main aspects of intelligent fashion techniques. These two surveys focus on either a special generative model or a particular application on human image generation. Sha et al. [148] conduct a broad review of deep person generation with an excessive large scope, encompassing face, pose, and garment-oriented generation tasks. In contrast to their approach, this survey is structured from the perspective of method paradigms rather than specific tasks. We introduce more comprehensive papers on the topic of human image generation. Specifically, we provide a comprehensive coverage of recent Transformer, NeRF, and diffusion model-based data-driven methods, knowledge-guided methods, and hybrid methods, addressing aspects that may be lacking in [148]. Liao et al. [91] propose another survey focusing on appearance and pose-guided human image generation. In comparison to [91], primarily focusing on GAN and diffusion model-based methods, our work offers more comprehensive method taxonomies and introduces a broader array of methods, datasets, evaluation metrics, and other pertinent factors. In summary, compared to previous surveys, we present a thorough review of human image generation, encompassing three-category generation methods, distinct taxonomies and pipelines, evaluation benchmarks, applications, and discussions of challenges and opportunities, aiming to provide researchers with a comprehensive overview of this topic.

The following parts of this survey are organized as follows. Firstly, we introduce the most representative methods of the three paradigms in human image generation, namely the data-driven methods in Section 2, the knowledge-guided methods in Section 3 and the hybrid methods in Section 4. Following that, we delve into the commonly used benchmark datasets, evaluation metrics, and conduct a qualitative and quantitative comparison of representative methods in Section 5. Section 6 provides an overview of the related applications of synthesized human images. Finally, we explore the challenges and potential research opportunities in Section 7.

## 2 DATA-DRIVEN METHODS ON HUMAN IMAGE GENERATION

As introduced in the preceding section, human image generation has emerged as a thriving research topic alongside the development of deep neural networks and generative models. In this section, we initially categorize diverse data-driven methods according to their foundational models, including **GAN**, **VAE**, **U-Net**, **NeRF**, **Transformer and attention mechanism**, and **diffusion model**. Additionally, beyond foundational models, we introduce a taxonomy for data-driven methods based on task settings that accommodate various application requirements, such as **pose-conditioned image transfer**, **garment-conditioned image transfer**, and **hybrid conditional image transfer**. Subsequent to the proposed taxonomies, we delve into a detailed discussion of the three main components in the overall pipeline of data-driven methods, encompassing **feature encoding**, **image generation**, and **loss functions**.

### 2.1 Method Taxonomy Based on Fundamental Models

In research on data-driven human image generation, a variety of deep neural networks and generative models are employed, each representing distinct technical approaches in this field. Consequently, this section initiates with the proposal of a method taxonomy grounded in fundamental models. For a concise overview, Table 1 summarizes the various fundamental models, their strengths and weaknesses, and representative methods with different feature encoding and image generation manners.

*2.1.1* **GAN-Based Methods.** GAN is initially proposed by Goodfellow et al. [49] to model the data distribution from training samples by training a generator and a discriminator to find the Nash equilibrium of a zero-sum game. The basic loss function of GAN can be formalized as follows.

$$\min_G \max_D \mathcal{L}_{GAN}(D,G) = \mathbb{E}_{(x \sim p_{data}(x))}[\log D(x)] + \mathbb{E}_{z \sim p_z(z)}[\log(1 - D(G(z)))], \quad (1)$$

where $x$ indicates a real image and $z$ indicates a noise signal which samples from a normal distribution. $D(\cdot)$ and $G(\cdot)$ represent the discriminator and generator, respectively.

Since then, researchers have dedicated significant efforts to developing various GAN variants [18, 77, 227]. For a more in-depth exploration of GAN models and their applications, researchers may refer to surveys [17, 99, 170, 209]. Given the potent capabilities of GAN models in fitting the target data distribution from extensive training samples, certain GAN variants, e.g., CycleGAN [227] and StyleGAN [77], have served as the foundational model for human image generation. Additionally, the AdaIN layer [63] has been employed to transfer the imaging style from the conditional human image to the target one.

Due to its first-mover advantage in generative neural networks, GAN-based human image generation has become the most widely adopted technical approach among data-driven methods. The GAN-based methods are suitable to generate novel images, while preserving the training data distribution with conditional information insertion. Consequently, various condition-guided methods have emerged for human image generation, e.g., pose-guided methods[1, 95, 133, 153, 155, 161, 194], text-guided methods [188, 217, 225, 228], and target image-guided methods [59, 83, 84, 181, 187]. Furthermore, the adversarial training framework also can be integrated with other fundamental models, e.g., VAE [211, 217], U-Net [1, 47, 69, 175], attention mechanism [85, 162, 229, 230], and NeRF [74, 182, 211].

Table 1. The Brief Characteristics of Data-Driven Methods Categorized by Fundamental Models.

| Method Taxonomy | Strengths | Weaknesses | Representative Methods |
|---|---|---|---|
| GAN-Based Methods | Most applied High fidelity High diversity Hight flexibility | Unstable training Mode collapse | Disentangled: [10, 87, 109, 113, 152, 185, 188] Self-Supervised: [47, 103, 110, 132, 163, 175] Flow: [31, 35, 59, 136, 137, 171, 220] Att.: [85, 136, 137, 162, 171, 187, 229, 230] |
| | | | Coarse-to-Fine: [34, 108, 181, 198, 217] S.-T.: [33, 83, 84, 134, 158, 203, 210, 228] Warping-Fusion: [28, 47, 69, 195, 197] |
| VAE-Based Methods | Stable training | Relatively low fidelity Blurry details | Disentangled: [75, 80, 205] Self-Supervised: [80] |
| | | | Segment-Texture: [211] |
| U-Net-Based Methods | Suitable for VITON High flexibility | Need warping process Low diversity Detail deficiency | Disentangled:[185] Self-Supervised: [157] Flow: [21, 48, 55, 154, 180, 192] Attention: [21] |
| | | | Coarse-to-Fine: [56, 217] Segment-Texture: [55, 106, 157, 178] W.-F.: [20, 21, 48, 69, 154, 167, 180, 192, 219] |
| Transformer & Attention-Based Methods | High fidelity High diversity Hight flexibility | High computation cost Large model scale | Disentangled: [15, 75, 135, 226] Self-Supervised: [40, 213] Flow: [8, 73, 107] |
| | | | Coarse-to-Fine: [40] Warping-Fusion: [8, 73, 196] |
| NeRF-Based Methods | Full view synthesis | Need serialized training data High computation cost Low diversity and flexibility | Self-Supervised: [72] |
| | | | Coarse-to-Fine: [142] Segment-Texture: [211] |
| Diffusion Model-Based Methods | Stable training High fidelity High diversity Hight flexibility | High computation cost Large model scale Slow inference | Disentangled: [15, 75, 80, 212] Flow: [80, 123] Attention: [57] |
| | | | Coarse-to-Fine: [150] Warping-Fusion: [89, 123] |

"Att.", "S.-T." and "W.-F." are short for "Attention", "Segment-Texture" and "Warping-Fusion", respectively.

2.1.2 **VAE-Based Methods**. The variational autoencoder (VAE) [79] is introduced as an unsupervised generative model, which has the capability to generate new images from random noise signals. Assuming that the input data follows a normal distribution, a noise signal $\epsilon$ sampled from the distribution of the input data is fed into the decoder to reconstruct the input images. Throughout the training process, a reconstruction loss and the Kullback-Leibler divergence (KL-divergence) are utilized as the supervision signals. Formally,

$$\mathcal{L}_{VAE}(x, \theta, \phi) = \mathbb{E}_{z \sim q_\phi(z|x)} \left[ \log p_\theta(x|z) \right] - D_{KL} \left[ q_\phi(z|x) \parallel p_\theta(z) \right], \quad (2)$$

where $q_\phi(\cdot)$ and $p_\theta(\cdot)$ indicate the operations of encoder and decoder, respectively. $D_{KL}(\cdot)$ indicates the KL-divergence between two distributions.

The VAE assumes that the data follows normal distributions and aims to minimize the KL-divergence between the real data distribution and the generated data distribution, which is challenging to achieve perfectly. Due to the averaging property of the distribution approximation, images synthesized by VAE tend to be relatively blurred compared to results from GAN. In general, VAE-generated images may exhibit more synthesized (unnatural) traits compared to models based on GAN. Consequently, only a few data-driven methods [42, 75, 205, 211] utilize VAE models as their backbones.

*2.1.3* **U-Net-Based Methods**. The U-Net architecture [140] is originally designed to address biomedical image segmentation challenges. U-Net introduces long skip connections to make the high-resolution features from the encoder path to combine and reuse with the upsampling decoder, so that multi-scale information can be fused for superior segmentation results.

Since U-Net is designed to learn a pixel-to-pixel mapping function and the skip connection operations can fuse both low-level and high-level features to complement missing information during the encoding-decoding process, this framework is widely employed in human image generation tasks [13, 21, 55, 106, 154, 157, 178, 180, 185].

*2.1.4* **Transformer and Attention-Based Methods**. Transformer [166] is a deep neural network with an encoder-decoder architecture, initially proposed for machine translation. The foundation of the Transformer network is the self-attention mechanism, adept at capturing the long-term dependencies of input samples. Transformer also adopts a parallel multi-head attention mechanism, which learns attentional information in different representation subspaces. Vision Transformer (ViT) [39] brings this architecture into CV research. It divides the input image into a series of fixed-size patches, which are then processed as sequenced tokens in a Transformer encoder, achieving impressive performance on image classification tasks. For more information on Transformer, researchers can refer to the surveys [54, 78].

Considering the effectiveness of Transformer and its attention mechanism in CV and NLP, researchers also adopt them in the field of human image generation. To better infer the regions of interest on human pose, shape, and appearance, some methods [15, 85, 136, 187, 196, 226, 230] implement attention mechanisms as core modules in model networks. Among these methods, attention mechanisms are combined with pose and appearance flow estimation methods to capture fine-grained details during the novel image generation process [8, 21, 107, 136, 137, 171]. Additionally, attention mechanisms are widely adopted in pose-guided human image generation pipelines to provide more precise features for pose transfer [135, 162, 187, 229, 230]. Furthermore, complete Transformer architectures can also be utilized as model backbones for human image generation [40, 73, 86, 213].

*2.1.5* **NeRF-Based Methods**. Recently, the neural radiance field (NeRF) model [114, 115] has been proposed for synthesizing novel view images of complex 3D scenes. The NeRF model represents a 3D scene as an implicit function that optimizes volume density and color at continuous locations based on a series of input images from different view angles. The implicit function is modeled by a MLP network, which takes 3D location coordinates $(x, y, z)$ and 2D viewing directions $(\theta, \phi)$ as inputs to produce color values $(r, g, b)$ and volume density $\sigma$. Following this, the volume rendering techniques are applied to composite the images of novel views. For more detailed information on NeRF, researchers can refer to the survey [45].

Given its proficiency in synthesizing novel view images of 3D scenes, NeRF is naturally applied in human image generation to create human image of different views. The HumanNeRF model [176] takes monocular video of a human as input and optimizes a canonical volumetric T-pose of the human, to render the subject from arbitrary novel views. Building on the foundation of HumanNeRF [176], several NeRF-based free-viewpoint human body rendering methods [72, 177, 206] have been introduced. Furthermore, the 3D-SGAN model [211] combines a generative NeRF [147] with a texture generator. The generative NeRF learns an implicit 3D representation of the human body and outputs 2D semantic segmentation masks, while the texture generator transforms these masks into a real image with human appearance texture. Before NeRF gains popularity, researchers attempt to synthesize novel view human images with a similar concept, representing 3D human bodies as implicit functions. The PIFu model [141] consists of a fully convolutional image encoder and an implicit function represented by an MLP network to reconstruct textured human surfaces.

PIFuHD [142] further provides higher resolution images through a coarse-to-fine procedure, serving as a pre-trained reconstructor in subsequent works on 3D human avatar generation [74, 182]. For brevity, these methods are also categorized as NeRF-based methods in this survey.

*2.1.6* **Diffusion Model-Based Methods**. Recently, diffusion models have emerged as an explosively captivating research topic in generative artificial intelligence. The core idea is to train a Markov diffusion chain to map input data to a noise distribution, which gradually adds Gaussian noise to input data, termed the forward processes. Then, the reverse process can be applied to convert noise the distribution back into the target distribution, i.e., generating target images from the Gaussian noise. The denoising diffusion probabilistic model (DDPM) [62] employs a U-Net-based diffusion model with residual blocks, self-attention mechanisms, and a simplified training objective. DDPM showcases the capability of the diffusion model to generate high-quality images. The remarkable latent diffusion model (LDM) [139] introduces a compressed lower-dimensional latent space for the diffusion process, delivering competitive performance in image synthesis, inpainting, and image super-resolution with limited computational resources. For more details of diffusion models, researchers can refer to the survey [199].

Given their impressive performance in image synthesis, diffusion models have found extensive application in human image generation. The person image diffusion model (PIDM) [15] introduces a pose-guided high-fidelity human image generation framework. It incorporates a texture diffusion block with cross-attention and disentangled guidance to establish correlations between appearance and pose information. The progressive conditional diffusion models (PCDMs) [150] are also crafted for pose-guided human image generation, employing a coarse-to-fine pipeline. Pose-constrained latent diffusion (PoCoLD) [57] goes a step further by incorporating DensePose [53] as the target pose condition within a U-Net-based latent diffusion model. The latent flow diffusion model (LFDM) [123] generates desired videos from images by synthesizing an optical flow sequence in the latent space. In addition to pose-guided methods, HumanDiffusion [212] introduces a text-driven garment style transfer approach. For the virtual try-on task, the keypoints guided inpainting (KGI) method [89] employs a diffusion model to progressively synthesize novel images based on target segmentation maps. Multimodal garment designer (MGD) [11] and LaDI-VTON [118] implement stable diffusion [139] in their methods, each with different prompts. The prompt-free diffusion model [189] can also be applied to virtual try-on tasks with specified garment conditions. Moreover, the Affordance-Insertion model [80] proposes the insertion of a target person into different backgrounds using the diffusion model.

## 2.2 Method Taxonomy Based on Task Settings

In addition to categorizing data-driven methods based on their fundamental models, this survey also introduces a method taxonomy based on different task settings. As illustrated in Figure 3, most literature predominantly falls into three task settings in human image generation: **pose-conditioned image transfer**, **garment-conditioned image transfer**, and **hybrid conditional image transfer**.

The pose-conditioned image transfer aims to alter the original person image according to a target pose. The pose guided person generation network (PG$^2$) [108], as a pioneering work, establishes the paradigm for pose-conditioned human image generation. Illustrated in Figure 4(a), the model takes a source image and a target pose as inputs, synthesizing a novel human image with the source appearance and the target pose. This task setting, introduced by the PG$^2$ model, has inspired numerous studies in recent years, including [85, 106, 110, 137, 161, 210, 220]. In these works, addressing the encoding of pose features from the target image has become a crucial challenge.

The garment-conditioned image transfer focuses on changing a person's garments using a target clothing image or a person image with target clothing. The first paradigm is introduced by Han et al.[56] in the context of VITON, as depicted in Figure 4(b). VITON is the pioneering data-driven human image generation method for garment image-conditioned virtual try-on. For methods focused on garment-conditioned virtual try-on [20, 28, 47, 48, 180, 192, 219], the target clothing images are essential to guide the generation process. A common pre-processing step involves extracting human body segmentation maps or clothing masks for garment-conditioned human image generation.

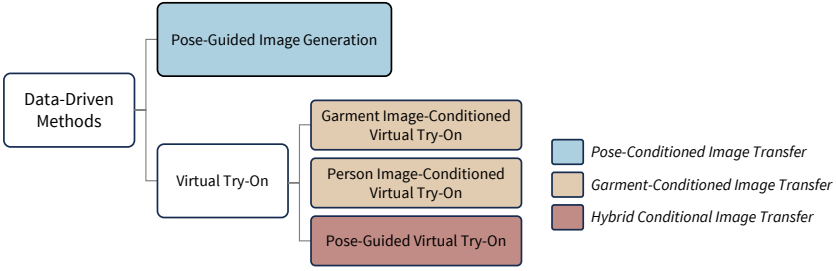

Fig. 3. Data-driven method taxonomy based on different task settings.

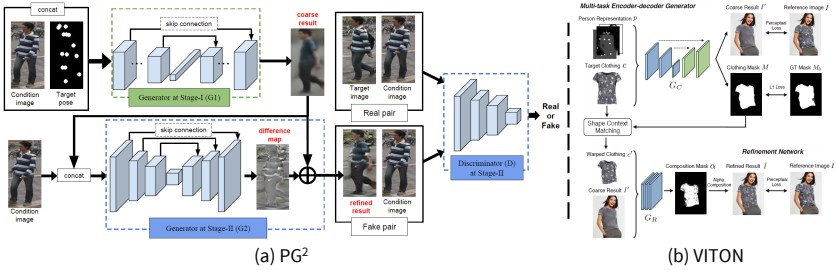

(a) PG$^2$     (b) VITON

Fig. 4. The overviews of the PG$^2$ model [108] and the VITON model [56].

The hybrid conditional image transfer involves changing both the garment and the pose simultaneously. These methods [21, 31, 34, 35, 187] utilize both a target pose representation and a target clothing or person image as inputs. These methods replace both clothing and pose in the source image with the corresponding target attributes while preserving other attributes (e.g., face and gender) from the source image. In addition to visual conditions provided by a target image, text descriptions can also serve as guidance for synthesizing person images [217, 225, 228].

## 2.3 Main Components in Data-Driven Methods

Figure 5 illustrates the overall pipeline of data-driven human image generation, consisting of three main components: **feature encoding**, **image generation**, and **loss functions**. In this section, we introduce the commonly used settings and methods in these components.

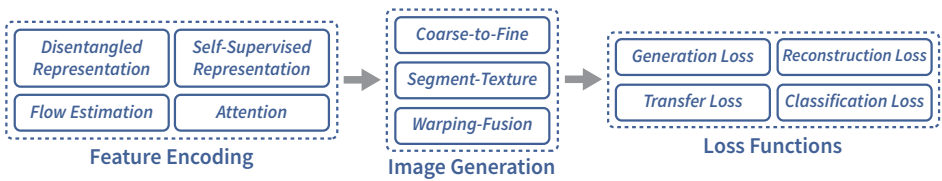

Fig. 5. The common flowchart of data-driven human image generation.

*2.3.1 **Feature Encoding**.* As depicted in Figure 5, prior to generating a novel person image, the input source image and the synthesis conditions (target images or textual attributes) must first be encoded into a latent feature space. In this space, transfer operations can be conducted to modify the source image in various styles. In this context, we discuss four main challenges related to feature encoding in human image generation: **disentangled representation learning**, **self-supervised representation learning**, **flow estimation**, and **attention mechanisms**.

***Disentangled Representation Learning***. Factorizing a human image into a set of disentangled latent variables, e.g., clothing attributes, body poses, and background styles, facilitates the manipulation of specific latent variables to generate a novel image. Consequently, learning disentangled representations becomes a crucial challenge for controllable human image synthesis.

The study [109] introduces a noteworthy model for acquiring disentangled representation from human images. Illustrated in Figure 6(a), the model explicitly decomposes a person image into three components: foreground, background, and pose, employing a multi-branch reconstruction network. This approach enables flexible generation of novel images by sampling from the disentangled features. Similar to the approach in [109], the synthesizing unseen poses model [10] employs body masks to distinguish between foreground and background elements. Additionally, it leverages masks for different body parts to disentangle the entire body, providing precise guidance for the generation process.

In addition to body masks, attribute-decomposed GAN [113] utilizes pixel-level human parsing to break down a human body into different body parts. Similarly, MUST-GAN [110] also employs human parsing for part-based disentangled representations. Moreover, DCTON [47] trains a mask prediction encoder-decoder network to disentangle the clothes region and skin region in a person image, using them as prior guidance for pixel warping and image generation.

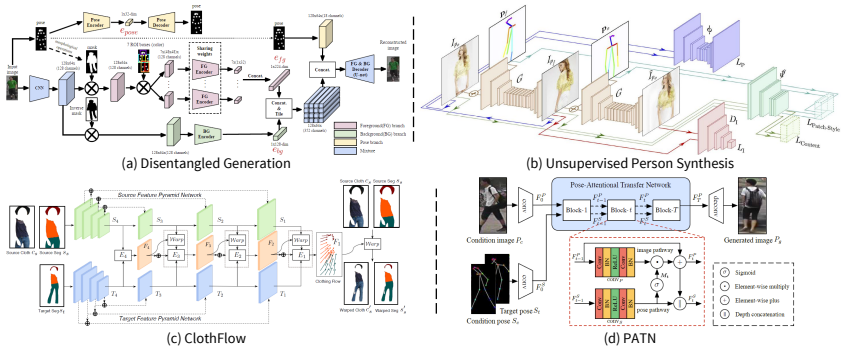

Fig. 6. The overviews of representative data-driven methods with different feature encoding strategies [55, 109, 132, 229].

***Self-Supervised Representation Learning***. In human image generation tasks, particularly virtual try-on, collecting large-scale paired image datasets poses a challenge. Therefore, the self-supervised representation learning framework proves to be suitable for this scenario.

Self-supervised learning methods leverage source images as reconstruction supervision to train the generation network. A notable example is the VITON model [56], employing a U-Net framework to recover source human images from extracted person representations, as depicted in Figure 4(b). The CP-VTON model [167] adopts a similar person representation method but skips VITON's coarse-to-fine procedure, resulting in finer synthesis with realistic details. The SwapNet [134] integrates a warping module and a texturing module. The dual-path warping module synthesizes the target clothing segmentation, while the texturing module is self-supervised to generate the source image with clothing segmentation and texture information. VTNFP [203] synthesizes warped clothing and body segmentation first, followed by the try-on synthesis module, incorporating the clothing image and person representation for the final image. DG-Net [222] introduces two source image and two feature reconstruction constraints for generating auxiliary images for person re-identification (Re-ID) tasks.

Within the self-supervised representation learning framework, cycle-consistency methods represent a unique approach. Originating from CycleGAN [227], these methods train generators to synthesize target images. Subsequently, the models use the synthesized target images as inputs to reverse-generate the source images, enabling the use of source images as supervision for training. Illustrated in Figure 6(b), the unsupervised person synthesis model [132] lays the foundation for the cycle-consistency reconstruction learning pipeline. Initially, the source image and target pose are combined to input the generator, synthesizing the human image with the target pose. Subsequently, the generator takes the synthesized image and the source pose as inputs to generate the original source image. Following that, the Semantic Parsing Transformation model [158] introduces a semantic generator to provide the body segmentation map as an intermediate step. The SwapGAN model [103] incorporates a body segmentation-conditioned generator in the second generation procedure. Moreover, in $C^2$GAN [163], three reconstruction cycles are set to train the model. Alongside the common image-to-image cycle, two key-point cycles are connected in the network to provide complementary information from different modalities.

*Flow Estimation*. Flow estimation is employed to learn dense correspondence in paired images, providing a description of deformations between source and target images in image synthesis. For instance, the ClothFlow model [55] focuses on estimating the appearance flow from the source clothes to the target persons. The appearance flow essentially indicates which pixels in the source clothing image could be applied in the target image. Subsequently, the model utilizes the estimated flow to warp the source clothes, facilitating the synthesis of the target image. An illustration of clothing appearance flow estimation in [55] is presented in Figure 6(c). In contrast to the appearance flow, the Fw-GAN model [35] introduces an optical flow guided fusion module, which is designed to warp past frames, enabling the synthesis of new frames for the pose-guided virtual try-on task. Similarly, the FashionMirror model [21] presents a skeleton flow extraction network, which learns feature-level optical flows from sequential poses, guiding the generation of target try-on clothing images. GFLA [137] utilizes a global flow field estimator to estimate the optical flow field from the source to the target, assisting in the image rendering process. Likewise, DiOr [31] incorporates the same global flow field estimator as GFLA [137] to warp the source person to the target poses. The PoseFlow [220] method introduces a pose flow estimator to guide the warping process, which aims to generate human image with novel poses.

*Attention Mechanism*. In the realm of human image synthesis, various attention models [85, 137, 229] have been deployed to augment the correlations of salient features during the transfer processes. The PATN model [229] incorporates a series of pose-attentional transfer blocks, wherein pose attention masks are employed to deduce the regions of interest. The framework of the PATN model and pose-attentional transfer block is illustrated in Figure 6(d). The PoNA model [85] adopts a framework similar to PATN [229] but updates the attention blocks with the non-local attention [172]. GFLA [137] utilizes local attention to spatially transform the source features and generate results. XingGAN [162] creates attention maps for both image and pose modalities, proposing a co-attention fusion module to merge appearance and shape features. Similarly, FashionMirror [21] employs co-attention to establish the relation between source images and target clothes for virtual try-on, avoiding explicit matching of body parts through time-consuming semantic parsing. Additionally, the STTEN model [187] implements a pose-guided high-frequency attention module to enhance the texture details of synthesized human images.

*2.3.2 Image Generation.* After manipulating human styles (pose, clothing, etc.) in the latent feature space, the next step is to decode the latent code to generate a realistic human image. As illustrated in Figure 5, the commonly used image generation pipeline can be divided into three categories: **coarse-to-fine generation**, **segment-texture generation**, and **warping-fusion generation**.

*Coarse-to-Fine Generation.* As the first pose-guided human image generation method, PG$^2$ [108] adopts a typical coarse-to-fine two-stage image generation procedure. As depicted in Figure 4(a), the model generates a coarse result based on the source image and target pose in the first stage. Subsequently, the model further refines the coarse result and synthesizes the final output image in the second stage.

The VariGANs model [217] synthesizes the coarse image with textual descriptions of viewpoints and further generates the high-resolution image with a U-Net encoder-decoder network. The VITON model [56] initially adopts a multi-task encoder-decoder generator to synthesize a coarse human image and the corresponding clothing mask. Then, the clothing details are warped to the mask of the coarse image. Following VITON, the MG-VTON model [34] synthesizes the coarse result using a Warp-GAN, and a refinement render is implemented to refine the coarse result.

*Segment-Texture Generation.* Segment-texture generation is another two-stage strategy for human image generation, where the model first generates the target segments of body parts based on the source images and synthesis conditions, and then renders the texture details on the segments of various body parts.

Zhu et al. [228] propose the first segment-texture pipeline in the FashionGAN model. As illustrated in Figure 7(a), FashionGAN utilizes text descriptions to provide attribute guidance for semantic parsing and texture rendering. Different from FashionGAN, Warping-GAN [33] uses target poses as guidance to generate segments of body parts. Similar to [33], SwapNet [134] consists of a warping module and a texturing module, where the warping module adopts a dual-path U-Net network to generate body segmentation maps with the target poses. Then, the texturing module generates the final images in a self-supervised manner. Lately, several studies [106, 158, 203, 210] also follow the segment-texture strategy for human image generation.

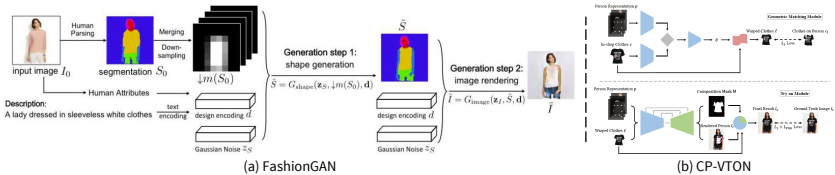

Fig. 7. The overviews of representative data-driven methods with different image generation strategies[167, 228].

*Warping-Fusion Generation.* The warping-fusion approach represents a distinctive strategy in human image generation, involving the initial warp of the target garment based on the specified pose. Subsequently, this contoured garment is seamlessly integrated with the target person to yield the final synthesized output.

The pioneering CP-VTON model [167] introduces the warping-fusion methodology. Illustrated in Figure 7(b), the model comprises a geometric matching module (GMM) and a try-on module, corresponding to the pivotal warping and fusion stages, respectively. Within the GMM, a thin-plate spline transformation is employed to generate the warped clothing, leveraging estimated spatial transformation parameters. The try-on Module utilizes a U-Net for predicting a composition mask and a rendered person image, both intricately fused with the warped clothes to produce the final try-on image. Building upon this foundation, DCTON [47] leverages a three-stream encoder-decoder network to generate target images, while VITON-HD [28] introduces an alignment-aware segment generator for synthesizing high-resolution virtual try-on images. M3D-VTON [219] further attempts to synthesize 3D virtual try-on mesh, by estimating the body depth map.

Diverging from traditional warping-fusion approaches [28, 47, 167, 195, 197], which necessitate explicit human parsing for mask extraction, the innovative WUTON model [69] introduces a novel paradigm with a student-teacher knowledge distillation framework. In a similar vein, the PF-AFN model [48] adopts the concept of parser-free methods. This inventive strategy combines the strengths of parser-based and parser-free models for enhanced performance in virtual try-on scenarios.

*2.3.3 Loss Functions.* In this section, we delve into the integral role of loss functions in data-driven methods, providing a comprehensive overview of their diverse applications. These loss functions, classified into four distinct categories based on their roles within the overall pipeline, encompass the **generation loss**, **transfer loss**, **reconstruction loss**, and **classification loss**.

*Generation Loss.* Within the realm of generation losses, pivotal components include the adversarial loss function as illustrated in Equation (1), and the KL-divergence as illustrated in Equation (2), which play a crucial role in GAN-based methods and VAE-based methods.

*Transfer Loss.* The family of transfer loss includes the style loss [46], the perceptual loss [76], and the contextual loss [111], etc.

The style loss function [46] is designed to guide models in synthesizing images that capture the same stylistic elements, such as colors, textures, and patterns, as the target images. This loss function relies on the Gram matrix to characterize the style of an image. Equation (3) illustrates the style loss for the $l$th layer feature map in a CNN network. Here, $I_g$ and $I_t$ represent the generated image and the target image, respectively. $G^l(x)$ denotes the Gram matrix of the image sample $x$ at the $l$th layer feature map $f(x)$, with dimensions $C \times H \times W$.

$$\mathcal{L}_{style}^l(I_g, I_t) = \|G^l(I_g) - G^l(I_t)\|_F^2, G^l(x)_{i,j} = \frac{1}{CHW} \sum_{h=1}^{H} \sum_{w=1}^{W} f^l(x)_{h,w,i} f^l(x)_{h,w,j}. \tag{3}$$

The perceptual loss function [76] is proposed to guide the generated image $I_g$ to exhibit similar CNN feature representations as the target image $I_t$. The formulation of the perceptual loss function at the $l$th layer of the CNN feature map is depicted in Equation (4). It's worth noting that due to its nature of reconstructing feature representations, this loss function is also referred to as content loss [132, 158], feature loss [134], and correspondence loss [210].

$$\mathcal{L}_{perc}^l(I_g, I_t) = \frac{1}{CHW} \|f^l(I_g) - f^l(I_t)\|_2^2. \tag{4}$$

The contextual loss function [111] is designed to assess the similarity between non-aligned data, allowing for spatial deformations between generated images and target images. In contrast to the strict requirement of feature spatial alignment in style loss and perceptual loss, the contextual loss takes a more flexible approach. Equation (5) formulates the contextual loss for the $l$th layer CNN feature map. $CX(\cdot)$ represents the similarity metric between different feature maps based on cosine similarity.

$$\mathcal{L}_{cont}(I_g, I_t) = -\log\left(CX\left(f^l(I_g), f^l(I_t)\right)\right). \tag{5}$$

*Reconstruction Loss.* The commonly employed reconstruction losses in person image generation include the L1 loss and the L2 loss. The L1 loss function is often utilized to guide the model in reconstructing target images on a pixel-to-pixel basis, aiding in reducing color distortions and expediting convergence speed [113]. Conversely, the L2 loss function typically serves as an auxiliary component in the training process of human image generation models, playing a role in tasks such as pose information reconstruction [109, 132, 158, 225] and optical flow learning [35, 137].

***Classification Loss***. The common classification loss, specifically the cross-entropy loss, plays a crucial role in human image generation. A large number of methods [35, 106, 134, 158, 197, 210, 220] employ the cross-entropy loss for training the human parsing or garment parsing modules within models. Beyond the cross-entropy loss, the focal loss [93] finds application in supervising pixel-wise segmentation within the VTNFP model [203]. In GFLA [137], cosine similarity is employed to measure the distance of features post the warping operation.

## 3 KNOWLEDGE-GUIDED METHODS ON HUMAN IMAGE GENERATION

Before the rise of generative models, early human image generation methods heavily rely on prior knowledge of human factors such as 3D body models, illumination conditions, and camera angles. In essence, the early methods of human image generation [22, 71, 129, 130, 138, 186, 200, 224], fall under the category of knowledge-guided methods. In this section, we initially introduce the representative fundamental models of knowledge-guided methods, namely SCAPE [7] and SMPL [105]. Subsequently, to facilitate a clear understanding of the diverse and intricate methods, we simplify and categorize the two primary pipelines employed in knowledge-guided human image generation: the **pixel warping pipeline** and the **virtual rendering pipeline**. For ease of reference, Table 2 provides a summary of representative knowledge-guided methods of different pipelines and fundamental models.

Table 2. Representative Knowledge-Guided Methods of Various Pipelines and Fundamental Models.

| Method Pipelines | Fundamental Models | Representative Methods |
| --- | --- | --- |
| **Pixel Warping Pipeline** | SCAPE [7] | Parametric Reshaping [224], MovieReshape [71] ReshapeData [130], Reshaping the Future [129] |
|  | - | Video-Based Characters [186] |
| **Virtual Rendering Pipeline** | SCAPE [7] | Synthesizing 3D Pose [22] |
|  | SMPL [105] | SURREAL [165], ClothCap [131], Octopus [3] Video Based Reconstruction [4], Pix2Surf [116] Multi-Garment Net [14] |
|  | Statistical Human [58] | Garment Recovery [200] |
|  | DeepDaz [191] | UltraPose [191] |
|  | - | MoCap-Guided [138], SOMAnet [12], PersonX [159] SyRI [9], RandPerson [173], UnrealPerson [215] |

### 3.1 Fundamental Models of Knowledge-Guided Methods

*3.1.1* ***Shape Completion and Animation for People (SCAPE)***. The SCAPE model [7], short for shape completion and animation for people, is specifically developed for constructing 3D human bodies with varying poses and shapes. This model comprises a pose deformation component and a shape deformation component. The pose deformation model effectively manages pose alterations by addressing rotations of rigid body parts and non-rigid deformations of the body independently. Firstly, the authors collect a pose dataset, encompassing 70 distinct poses of a specific individual, and a body shape dataset, featuring 37 different individuals in similar poses. A template mesh is chosen from the pose dataset, while the remaining meshes serve as instance meshes. The SPACE model employs the model mesh to generate the required 3D human body mesh using rigid part rotations $R$, non-rigid transformation $Q$, and shape deformation $S$. The 3D human body mesh $Y$ corresponding to given $R$, $Q$, and $S$ is constructed based on the model mesh $X$, represented as $Y = RSQX$. This enables the SPACE model to generate 3D human body meshes of various poses and shapes by manipulating $R$, $Q$, and $S$.

*3.1.2* **Skinned Multi-Person Linear (SMPL).** As its name implies, the SMPL model [105] is specifically designed to generate skinned human body model linearly for different individuals. Similar to SCAPE, SMPL also decomposes the 3D human body into identity-dependent shape and non-rigid pose-dependent shape, corresponding to shape and pose in SCAPE. SMPL is a statistical-based model that learns a mapping from the model parameter space to the vertex coordinate space, denoted as $M(\vec{\beta}, \vec{\theta}; \Phi) : \mathbb{R}^{|\vec{\beta}| \times |\vec{\theta}|} \mapsto \mathbb{R}^{3N}$. Here, $\vec{\beta}$ and $\vec{\theta}$ represent the shape parameters and pose parameters. $\Phi$ represents the model parameters that need to be learned. SMPL defines a body mesh with $N = 6890$ vertices and $K = 23$ joints. The basic 3D human body model is characterized by a mean template shape $\bar{T}$ with the rest pose $\vec{\theta}^*$ and a set of blend weights $\mathcal{W}$. Subsequently, shape blend shapes, pose blend shapes, and dynamic blend shapes are additively applied to $\bar{T}$. The 3D human body, in various shapes and poses, can be animated after obtaining the model parameters $\Phi$ by manipulating $\vec{\beta}$ and $\vec{\theta}$.

## 3.2 Pixel Warping Pipeline

The pixel warping pipeline is a classic approach in knowledge-guided human image generation methods. Methods such as [71, 129, 130, 186, 224] are characterized by their attempt to warp existing individuals in images to synthesize new ones. The general framework of pixel warping pipeline is illustrated in Figure 8.

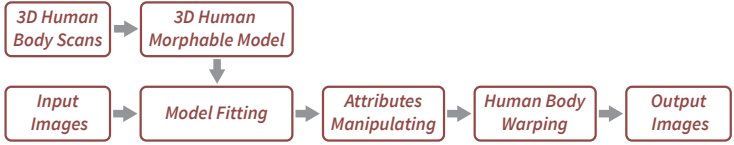

Fig. 8. The pixel warping pipeline of knowledge-guided methods.

In the pixel warping pipeline, a morphable 3D body model, such as SCAPE [7] and SMPL [105], is initially constructed using 3D laser scans of human bodies. Following that, the morphable 3D body model is fitted to the human bodies in the input images, as depicted in the *Model Fitting* block in Figure 8. Subsequently, attribute parameters can be manipulated to alter the shape or pose of the 3D morphable model, generating new 2D projections of human bodies. Finally, the human bodies in 2D projections undergo warping using methods such as moving least squares (MLS) [146] or other image deformation techniques [149]. The warped human bodies are then composited with various backgrounds to produce the final output images.

In the *3D Human Body Scans* and *3D Human Morphable Model* blocks of this pipeline, methods typically leverage existing data and pre-trained 3D human body models. The primary emphasis in methods following the pixel warping pipeline lies in enhancing the *Model Fitting*, *Attributes Manipulating*, and *Human Body Warping* blocks to address diverse tasks. Parametric Reshaping [224], for instance, introduces a user-interactive view-dependent model fitting method and a body-aware image warping approach to handle imperfect fitting, occlusions, and variations in clothes and poses. MovieReshape [71] employs a marker-less motion capture approach to fit pose and shape parameters of the SCAPE model to an actor's silhouette in videos. Subsequently, an MLS-based warping approach is utilized to alter the actor's shape. In contrast to fitting 3D human body models directly to individuals in images, video-based characters [186] retrieves frames from its database based on query skeletons. The vertex correspondence between the query mesh and the database mesh guides the warping process. The ReshapeData method [130] adopts the MovieReshape technique to generate synthetic human images for auxiliary training in the pedestrian detection task. Furthermore, depicted in Figure 9(a), the reshaping the future model [129] fits the human body model to annotated 3D poses by optimizing Euclidean distances between a set of 3D joint positions.

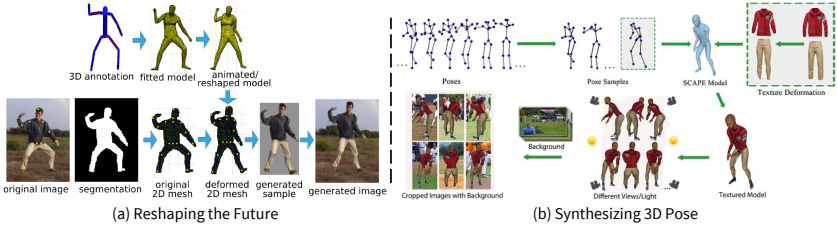

Fig. 9. The overviews of the representative knowledge-guided methods [22, 129].

In essence, methods within the pixel warping pipeline tend to focus on editing original human images rather than generating entirely new ones. Additionally, these approaches often necessitate substantial user interaction, such as providing skeleton information or aiding the model in segmenting person silhouettes during the synthesis of novel human images. The applications of pixel warping methods extend to image editing [224], movie and animation production [71, 186], and assisting the model training for other computer vision tasks [129, 130].

### 3.3 Virtual Rendering Pipeline

In addition to pixel warping, virtual rendering represents another common pipeline in knowledge-guided human image generation methods. The approaches within this pipeline [9, 12, 14, 22, 30, 131, 138, 159, 165, 173, 191, 200, 208, 215] aim to directly generate virtual human beings using 3D human body models. The framework of the virtual rendering pipeline is depicted in Figure 10.

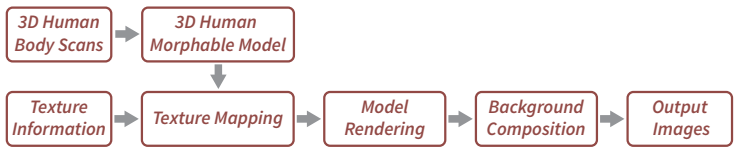

Fig. 10. The virtual rendering pipeline of knowledge-guided methods.

In contrast to the pixel warping pipeline, the methods in the virtual rendering pipeline do not rely on warping existing individuals in images. Instead, they follow a series of steps. Firstly, 3D human body models, such as SCAPE and SMPL, are constructed by learning from 3D human body scanning data. Subsequently, texture information, such as facial or clothing images, is mapped onto the 3D body models to obtain textured human bodies in various poses, i.e., the *Texture Mapping* block in Figure 10. Following that, the 3D human bodies, along with camera viewpoint and lighting information, are rendered to enhance the generation results in *Model Rendering*. Finally, in the *Background Composition* block, background images are merged with the 3D human bodies to produce the final synthesized human images.

The synthesizing 3D pose model [22] stands as a representative virtual rendering method, illustrated in Figure 9(b). This method generates over 5 million images for the 3D pose estimation task through body pose space modeling, clothing texture transfer, image rendering, and background composition. In the MoCap-Guided model [138], a synthesis engine is developed to combine body parts from a library of 3D motion capture (MoCap) data based on queried poses. The ClothCap model [131] involves aligning SMPL human body models with scan sequences, followed by garment segmentation. Subsequently, the garments can be retargeted to novel human bodies, making it applicable in virtual try-on scenarios, although with limitations on garment categories. The authors of the SURREAL dataset [165] pose the SMPL 3D human bodies guided by the MoCap data and render the images with background images, body texture maps, lighting and camera positions to synthesize the training data for human body part segmentation and depth estimation. The human appearance transfer model [208] employs a SMPL model to generate triangle meshes of monocular person images. Subsequently, the method transfers the appearance from a source image to a target image while maintaining the pose of the person in the target image. In multi-garment net [14], the authors compile a 3D garment wardrobe and transfer them to SMPL human body models with various shapes and poses. More recently, several methods [9, 12, 159, 173, 191, 215] have also explored the utilization of commercial software or game engines, e.g., *Adobe Fuse CC*, *MakeHuman*, *Unity3D*, and *Daz*, to generate large-scale human images with diverse human bodies, garments, and backgrounds.

The synthesized human images produced by the virtual rendering pipeline find extensive applications as training data in various computer vision tasks, including 3D pose estimation [22, 138], person segmentation [165], and person Re-ID [159, 173], among others. However, it's worth noting that images generated by virtual rendering methods are more prone to being recognized as synthetic compared to the results of pixel warping methods, as illustrated in Figure 12.

## 4  HYBRID METHODS

In Section 2 and Section 3, we have delved into the distinctions between data-driven and knowledge-guided methods. These two technical approaches exhibit significant differences. Knowledge-guided methods rely on 3D human body models, enabling the generation of individuals that can be flexibly modified. Conversely, data-driven methods leverage neural networks and generative models, facilitating effective model parameter training with large-scale data for efficient image generation. Recognizing the unique strengths of each approach, researchers have proposed a fusion of knowledge-guided and data-driven methods, referred to as hybrid methods in this survey. This integration aims to harness the benefits of both techniques, offering a more comprehensive and versatile approach to human image generation.

The defining feature of hybrid methods lies in their integration of 3D human body models with various neural networks. In essence, hybrid methods can be viewed as data-driven methods complemented by 3D human-related auxiliary knowledge, such as that from SCAPE [7], SMPL [105], or the UV maps provided by SMPL [105] and DensePose [53].

The key statistics of hybrid methods, encompassing fundamental models, and representative methods of various feature encoding strategies and image generation pipelines, are presented in Table 3. The fundamental model architectures of hybrid methods integrate both trainable neural networks and 3D human body models. Serving as a hybrid of data-driven and knowledge-guided methods, these approaches exhibit main characteristics that combine the strengths of both methodologies. As indicated in Table 3, hybrid methods share similar feature encoding strategies (disentangled representation, self-supervised representation, flow estimation, and attention mechanism) and image generation pipelines (coarse-to-fine, segment-texture, and warping-fusion) with data-driven methods. Therefore, we forego revisiting the frameworks and pipelines, previously discussed in Section 2 and Section 3, to concentrate on the hybrid methods.

As evident from Table 3, the methods relying on GAN models also represent a significant portion of hybrid methods, mirroring the prevalence observed in data-driven methods. A significant portion of these methods utilize SMPL [105] or its derivative models [16, 126] as foundational 3D human prior knowledge, categorized as GAN-SMPL-based methods in this survey [37, 38, 50, 179, 193, 201, 204, 216]. Some of them integrate DensePose [53] to provide ample information for pose-guided generation, categorized as GAN-DensePose-based methods [2, 100, 121, 145]. Additionally, the other 3D skeletal pose formulation methods, such as VNect [112], can also be implemented in hybrid methods [98].

Table 3. Representative Hybrid Methods of Various Fundamental Models.

| Fundamental Neural Networks | Fundamental 3D Models | Representative Methods |
|---|---|---|
| GAN | SMPL | Disentangled: [44, 101, 102, 193, 201]<br>Self-Supervised: wFlow [37]<br>FLow: [37, 88, 101, 102]<br>Attention: Attentional Liquid Warping GAN [102] |
| | | Coarse-to-Fine: Part-Based Representation[179], AG3D [38]<br>Segment-Texture: [37, 82, 193]<br>Warping-Fusion: DMM [204], TightCap [24]<br>Texture-Rendering: [5, 50, 82, 202, 216] |
| | DensePose | Disentangled: Outfit-VITON [121] |
| | | Coarse-to-Fine: Dense Pose Transfer [122]<br>Segment-Texture: Outfit-VITON [121]<br>Texture-Rendering: [2, 51, 100, 122, 144, 145, 207] |
| VAE | SMPL | Disentangled: UnitedHuman [44] |
| | | Segment-Texture: ClothNet [81] |
| | DensePose | Texture-Rendering: ReAVAE [19], HumanGAN [144] |
| U-Net | SMPL | Coarse-to-Fine: RANA [68]<br>Texture-Rendering: Re-ID Texture Generation [169], ARCH [66] |
| | DensePose | FLow: ZFlow [29] |
| | | Segment-Texture: ZFlow [29]<br>Warping-Fusion: ZFlow [29]<br>Texture-Rendering: Few-Shot Motion Transfer [65] |
| Transformer | SMPL | Self-Supervised: Texformer [190]<br>FLow: Texformer [190]<br>Attention: Texformer [190] |
| | | Texture-Rendering: Texformer [190] |
| | DensePose | Disentangled: $PT^2$ [86]<br>Self-Supervised: $PT^2$ [86]<br>Attention: $PT^2$ [86] |
| NeRF | SMPL | Disentangled: CustomHumans [61]<br>Self-Supervised: iVS-Net [36] |
| | | Texture-Rendering: Neural Actor [97], iVS-Net [36] |
| Diffusion | SMPL | Disentangled: UPGPT [27]<br>Self-Supervised: UPGPT [27] |
| | | Texture-Rendering: DINAR [160] |

Similar to data-driven methods, relative few hybrid methods utilize VAE as their fundamental generative neural networks [19, 44, 81, 144]. However, the ClothNet model [81] stands out as the pioneer hybrid human image generation method, incorporating the 3D human body model into the deep generation network.

The U-Net architecture plays a pivotal role in hybrid methods, serving as a key component integrated with SMPL [66, 68, 169] and DensePose [29, 65]. These architectures are utilized to generate diverse outputs, including various texture maps [29, 65, 66, 68, 169], normal maps [66, 68], semantic segmentation masks and UV maps [29, 65]. Moreover, U-Net architectures are extensively utilized as pixel-to-pixel mapping modules across various methodologies. They are integrated into GAN-SMPL-based methods [5, 24, 44, 88, 101, 102, 201], GAN-DensePose-based methods [51, 100, 144, 145], and VAE-SMPL-based models [44, 81].

In the realm of Transformer networks and attention mechanisms, Liu et al. [102] integrate attention mechanism into the liquid warping GAN model [101] to enhance feature fusion more effectively. On the other hand, Texformer [190] employs a Transformer network to estimate 3D human texture from a single image, leveraging attention mechanisms to extract global information and finer details. In the latest $PT^2$ model [86], cross-attention modules are employed to align texture features with target pose features for novel pose synthesis.

In Section 2.1.5, we introduce that NeRF is able to represents the color values and volume density of a 3D human scene with an implicit function, which is facilitated by a MLP network. It is feasible to integrate NeRF and implicit functions with parametric 3D human body models for novel view synthesis. Neural Actor [97] leads the training of NeRF models with SMPL guidance, enabling realistic free-view synthesis of human bodies. Subsequent works [23, 36, 127] leverage NeRF to predict the color and density of 3D human bodies. The ELICIT model [64] incorporates the SMPL model for 3D body geometry prior and CLIP features for clothing semantic prior to train HumanNeRF [176]. Furthermore, ActorsNeRF [120] aligns diverse human subjects in a category-level canonical space to generate novel actors with novel poses. Additionally, models like neural body [128], CustomHumans [61], ICON [184], and ECON [183] utilize SMPL for 3D human body modeling and MLP-modeled implicit functions for predicting color and density, enabling different view synthesis capabilities.

Recently, researchers have integrated diffusion models with the SMPL model for hybrid human image generation [27, 32, 57, 160]. The UPGPT model [27] introduces a multi-modal diffusion model that leverages text, pose, and visual prompts to guide the generation of novel human images. The SMPL model contributes pose parameters for the multi-modal diffusion process. Cloth2Body [32] proposes an end-to-end framework capable of accurately estimating 3D body mesh parameters, including pose and shape, from a 2D clothing image. DINAR [160] utilizes a latent diffusion model [139] to restore texture features, which are then combined with the SPML-X body model [126] to generate human avatars.

In terms of diverse feature encoding strategies, BodyGAN [193], UnitedHuman [44], and Outfit-VITON [121] utilize disentangled representation learning to synthesize different parts of the human body. 3DHumanGAN [201] employs two mapping networks to disentangle pose and appearance generation. Liquid warping GAN [101, 102] disentangles foreground and background contents through distinct flow pathways.

Similar to data-driven methods, appearance flow finds wide application in hybrid methods [37, 88, 101, 102]. The dense appearance flow model [88], for example, estimates dense 3D appearance flow to guide pose transfer. This model fits the SMPL 3D model to both the source and target poses, and then computes dense appearance flow to train the flow regression module. Additionally, the wFlow model [37] integrates the respective strengths of 2D pixel flow and 3D vertex flow within a self-supervised training scheme for synthesizing dancing videos. The ZFlow model [29] proposes gated appearance flow, which predicts appearance flow across multiple scales and aggregates them with a gating mechanism. The UV maps extracted from DensePose [53] serve as geometric priors in the ZFlow model.

As depicted in Table 3, certain hybrid methods [29, 81, 121, 122, 219] share common image generation frameworks with data-driven methods, e.g., coarse-to-fine generation, segment-texture generation, and warping-fusion generation. However, other hybrid methods exhibit a unique human image generation framework, termed **texture-rendering generation** in this survey [2, 50, 100, 144, 151, 160, 218] The prominent characteristic of the texture-rendering framework is its initial extraction of the UV texture-map. Subsequently, the texture information is either rendered onto 3D human bodies or warped to generate human images with novel poses or appearances.

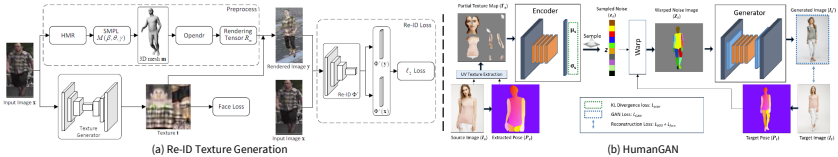

Fig. 11. The overviews of the representative hybrid methods [144, 169].

The Re-ID texture generation model [169] delineates a straightforward pipeline for texture-rendering generation, illustrated in Figure 11(a). This model generates the rendered image with the source person surface texture and a 3D SMPL mesh. Subsequently, a Re-ID model evaluates the similarity between the source and synthesized images, ensuring that the generated person images are suitable for the specific Re-ID task. The texture inpainting model [51] leverages the DensePose model [53] to convert poses into SMPL coordinate format. Initially, the model generates inpainted coordinate texture maps and RGB color texture maps, warping them based on the target pose. In the subsequent stage, the model utilizes feature maps from the first stage as inputs to synthesize the final person image with a U-Net generator. The dense pose transfer model [122] concurrently implements coarse-to-fine and texture-rendering generation pipelines. Initially, the model synthesizes a coarse target image conditioned on the target DensePose. Subsequently, it employs an inpainting autoencoder to generate the target texture view using the UV texture map as an intermediary. Finally, the coarse person image and the texture image are blended to produce the final result. In neural re-rendering [145], pose and shape are represented using the SMPL model, and partial UV texture maps are extracted to generate the full UV map with a U-Net based network. Then, the full UV map and the target pose serve as sources to synthesize the final photorealistic image using another generator network. In Figure 11(b), the HumanGAN model [144] first extracts the SMPL UV texture using DensePose. Then, a VAE-based encoder learns part-based latent vectors, achieving a disentangled representation for the body. SPATT [100] also utilizes DensePose to establish SMPL-format textures for human bodies. Similar to Neural Re-Rendering [145] and Texture Inpainting [51], the SPATT model initially generates the full UV texture for the input person. Then, the source image, full UV map, and target pose are blended to generate the output results.

## 5 DATASETS AND EVALUATION METRICS

### 5.1 Human Image Generation Datasets

***Market-1501***. The Market-1501 dataset [221] serves as a representative dataset in person Re-ID research. This dataset contains 32,668 images of 1,501 identities in different surveillance views. While the image resolution is low (128×64), Market-1501 provides large-scale image pairs featuring the same individual across various camera views, which makes it particularly valuable for pose-guided generation under video surveillance scenarios.

***DeepFashion***. The DeepFashion dataset [104] stands out as a detailed annotated large-scale clothing dataset, containing over 800,000 images sourced from various contexts such as shop displays, consumer-taken photos, and street snapshots. In particular, the in-shop clothes retrieval benchmark comprises 52,712 high-resolution images representing 7,982 distinct clothing items. This benchmark comprises image pairs featuring individuals in different poses but dressed in the same clothes. Researchers leverage this benchmark to train human image generation models aimed at altering clothing or synthesizing individuals in various poses.

*Zalando*. The Zalando dataset [56] comprises 16,253 pairs of frontal-view images of women and top clothing. The training and test set consists 14,221 pairs and 2,032 pairs respectively. During testing, the clothing images are randomly shuffled to simulate a real virtual try-on scenario.

***Chictopia10K***. The Chictopia10K dataset [90] comprises 10,000 images sourced from the Chictopia fashion website. It offers 18-class fine-grained semantic segmentation of various clothing categories and body parts. The ClothNet model [81] extends Chictopia10K by incorporating pose and shape features through the fitting of the SMPL model to the images.

***Human3.6M***. The large-scale Human3.6M dataset [67] comprises 3,578,080 3D human poses alongside corresponding images, enacted by 5 female and 6 male actors from 4 viewpoints. Featuring 17 distinct scenarios and human activities, e.g., eating, taking photos, sitting, and walking, the dataset also offers precise 3D body scans for all actors. Given its comprehensive nature, Human3.6M serves as an ideal resource for knowledge-guided and hybrid human image generation.

***Multi-View Clothing (MVC)***. The Multi-View Clothing (MVC) dataset [96] is designed for clothing retrieval and style recognition, furnishing 161,260 annotated images paired with 264-dimension attribute labels. Sourced from online shopping websites, the dataset encompasses images across four distinct views: front, back, left, and right. The MVC dataset proves particularly conducive to virtual try-on tasks.

***MPV and MPV-3D***. The MPV dataset [34] comprises 35,687 human images and 13,524 clothing images, showcasing individuals in various poses. The authors extract 62,780 three-tuples of the same person in the same clothes but with different poses. This dataset is instrumental for tasks related to pose-guided image generation and virtual try-on applications. Additionally, the MPV-3D dataset [219] is proposed based on MPV dataset for monocular-to-3D try-on applications. With 6,566 clothing-person image pairs, each image is coupled with front and back depth maps derived from the PIFuHD model [142].

***UltraPose***. UltraPose[191] represents a synthetic 3D human dataset generated by the DeepDaz model [191]. It comprises 500,000 individuals with a total of 1.3 billion corresponding point annotations on human body surfaces. On average, each person in the dataset is associated with 2,600 dense annotation points. This rich annotation density renders UltraPose particularly well-suited for training models aimed at pose-guided human image generation.

***Dance50k***. The Dance50k dataset [37] is proposed for real-world garment transfer task. This large-scale video dataset comprises 50,000 sequences of people dancing, featuring diverse garments. It serves as a valuable resource for human-centric research, e.g., virtual try-on, pose transfer, and human video synthesis.

***VVT***. The VVT dataset [35] is specially constructed for the video virtual try-on task. It contains 791 videos of fashion model catwalk with white color backgrounds, in which 661 videos for training and 130 videos for testing. The dataset also provides 791 person images and 791 clothes images associated with each video, making it suitable for the hybrid conditional image transfer task.

***RenderPeople***. RenderPeople is a prominent online library known for its extensive collection of high-resolution 3D human models. RenderPeople offers more than 4,500 individual 3D human models covering various topics, ethnicities and age groups. It is a valuable resource for high-resolution 3D human image generation, as evidenced by the works [141, 142, 183, 184].

***Thuman***. Thuman [223] emerges as a 3D real-world human model dataset designed for singe-image 3D human reconstruction [92, 168]. It contains 7,000 data items with approximately 230 distinct clothing variations under random poses. Each data item contains a textured surface mesh, an RGB-D image sourced from Kinect, and a corresponding SMPL model. This rich repository of data makes Thuman suitable for 3D pose-guided human image generation.

*VITON-HD*. The VITON-HD dataset [28] comprises 13,679 pairs of frontal-view woman images with corresponding top clothing items, at a resolution of $1024 \times 768$. The dataset is partitioned into a training set consisting of 11,647 image pairs and a test set containing 2,032 image pairs.

*Dress Code*. Dress Code [119] offers a comprehensive collection tailored for virtual try-on. It contains 53,795 pairs of front-view person-garment images, at a resolution of $1024 \times 768$. The dataset categorizes the image pairs into three groups: 15,366 pairs of upper-body garments, 8,951 pairs of lower-body garments, and 29,478 pairs of dresses. To facilitate further research, it provides human body key points, DensePose feature maps, and segmentation masks for each image pair.

*Stylish-Humans-HQ (SHHQ)*. The Stylish-Humans-HQ (SHHQ) dataset [43] is designed for unconditional human generation. Developed with the specific goal of training a StyleGAN-based human generation model, the SHHQ dataset offers a data-centric approach to model refinement. It contains 231,176 human images with $1024 \times 512$ mean resolution and provides labels of clothing types and textures, which is suitable for data-driven methods on various human generation and editing tasks.

*DNA-Rendering*. The DNA-Rendering dataset [25] is a large-scale human dataset with high-quality multi-view images and videos from various actors. The authors capture images from 500 actors employing 60 RGB cameras and 8 depth sensors, achieving resolutions of $4096 \times 3000$ pixels at 15 frames per second (FPS). Each actor showcases at least nine distinct motions with three different outfits, resulting in a dataset encompassing 5,000 videos with 67.5 million frames.

## 5.2 Human Image Generation Evaluation Metrics

*Manual Rating*. As a qualitative evaluation metric, manual rating requires substantial human involvement and is subject to inherent biases. However, assessing generated images through manual rating remains one of the most intuitive methods in evaluating different models. For manual rating evaluations, metrics such as **R2G**, **G2R**, and **Jab** can be employed [15, 226, 229]. R2G represents the percentage of real images classified as generated images by the evaluation participants. Conversely, G2R denotes the percentage of generated images classified as real images. Jab indicates the percentage of images judged as the best among all comparative models.

*Model Promotion Comparison*. As suggested by previous work [164], the evaluation of generative models should directly align with the applications or tasks for which they are designed. Consequently, an effective and equitable evaluation metric involves integrating the generated images into the training of corresponding CV tasks and assessing the models' advancements. Notably, tasks like person Re-ID [94, 108, 173] and 3D pose estimation [22, 138, 165] exemplify this approach, leveraging human image generation methods to synthesize supplementary training data.

*Inception Score (IS)*. The inception score [143] stands out as a popular quantitative evaluation metric in human image generation. The inception score leverages an inception network to classify the generated images $x$. As demonstrated in Equation (6), the calculation of the inception score is based on a KL-divergence. It comprises two primary components: the posterior distribution $p(y \mid x^i)$ and the distribution of classification labels $\hat{p}(y)$, where $x$ denotes the generated image samples and $y$ indicates their respective labels. Through the calculation of KL-divergence, the inception score tends to be large when a human image generation model excels in both quality and diversity.

$$IS(G) = \exp\left(\frac{1}{N}\sum_{i=1}^{N} D_{KL}\left(p(y \mid x^i) \parallel \hat{p}(y)\right)\right). \tag{6}$$

***Structural Similarity (SSIM)***. The structural similarity index (SSIM) [174] is another widely used quantitative evaluation metric in human image generation. It serves as a measure of similarity between the generated images and reference images, taking into account factors such as luminance, contrast and structure, formalized as Equation (7). In the equation, $x$ and $y$ represent the generated image and the reference image, respectively. $\alpha$, $\beta$, and $\gamma$ are typically set to 1. $\mu$ and $\sigma$ denote the mean and standard deviation of the respective images, while $\sigma_{xy}$ represents the correlation coefficient between $x$ and $y$. $C_1$, $C_2$, and $C_3$ are predefined constants. $I(x,y)$, $C(x,y)$, and $S(x,y)$ assess the similarity of luminance, contrast, and structure between the images, respectively.

$$SSIM(x,y) = I(x,y)^\alpha C(x,y)^\beta S(x,y)^\gamma, \text{where } I(x,y) = \frac{2\mu_x\mu_y + C_1}{\mu_x^2 + \mu_y^2 + C_1}, C(x,y) = \frac{2\sigma_x\sigma_y + C_2}{\sigma_x^2 + \sigma_y^2 + C_2}, S(x,y) = \frac{\sigma_{xy} + C_3}{\sigma_x\sigma_y + C_3}. \quad (7)$$

***Fréchet Inception Distance (FID)***. The Fréchet inception distance (FID) [60] serves as a metric for assessing the quality of images synthesized by generative models. FID operates under the assumption that both real and generated images conform to Gaussian distributions and quantifies the Fréchet distance between these distributions. Like the IS, FID leverages a pre-trained inception network to extract features from both real and generated images. FID can be regarded as an improvement over IS, which merely focusing on evaluating the distribution of synthesized images. The multidimensional Gaussian distributions representing the generated and real images are denoted as $\mathcal{N}(\mu_x, \Sigma_x)$ and $\mathcal{N}(\mu_y, \Sigma_y)$, respectively. Then, FID is calculated as follows.

$$FID(x,y) = \|\mu_x - \mu_y\|_2^2 + \text{tr}\left(\Sigma_x + \Sigma_y - 2\sqrt{\Sigma_x\Sigma_y}\right). \quad (8)$$

***Learned Perceptual Image Patch Similarity (LPIPS)***. LPIPS [214] is proposed to quantify the perceptual distance between pairs of images. It measures the distance between feature representations extracted from corresponding patches of two images by a CNN model. The LPIPS between images $x$ and $y$ can be expressed as follows.

$$LPIPS(x,y) = \frac{1}{N}\sum_{i=1}^{N} d\left(F_i(x), F_i(y)\right), \quad (9)$$

where $N$ represents the number of image patches, and $F(\cdot)$ denotes the image patch feature extracted by the CNN model. The distance function $d(\cdot)$ can implement various metrics, e.g., Euclidean distance and cosine similarity. LPIPS aims to imitate the perceptual capability of human vision rather than simply computing pixel-wise differences.

***Peak Signal-to-Noise Ratio (PSNR) and Average Matched Points (aMP)***. The peak signal-to-noise ratio (PSNR) is a traditional metric used in image and signal processing. It measures the similarity between the original image and the generated or reconstructed ones, as demonstrated in Equation (10). $MSE(x,y)$ calculates the average squared error between images $x$ and $y$, where the images have dimensions $m$ and $n$. $MAX$ represents the maximum possible pixel value in the image, which is typically 255 for an 8-bit image. A higher PSNR value indicates better generation or reconstruction quality.

$$PSNR(x,y) = 10 \cdot \log_{10}\left(\frac{MAX^2}{MSE(x,y)}\right), MSE(x,y) = \frac{1}{mn}\sum_{i=1}^{m}\sum_{j=1}^{n}(x(i,j) - y(i,j))^2. \quad (10)$$

The average matched points (aMP) is a traditional metric for image registration, which quantifies the average number of matched points between a reference image and a transformed image. Zhang et al. [211] propose using aMP to evaluate the 3D-view consistency of synthesized images.

***Chamfer Distance and Point-to-Surface (P2S) Distance***. The Chamfer distance is a commonly used metric for assessing the similarity between two point sets in 3D space. It computes the average distance from each point in one set to its nearest neighbor in the other set. For two point sets, $X$ and $Y$, the Chamfer distance is formulated as Equation (11). $x$ and $y$ represent the data points in sets $X$ and $Y$, respectively. $|X|$ and $|Y|$ denote the number of data points in each set.

$$CD(X, Y) = \frac{1}{|X|} \sum_{x \in X} \min_{y \in Y} \|x - y\|_2^2 + \frac{1}{|Y|} \sum_{y \in Y} \min_{x \in X} \|y - x\|_2^2. \quad (11)$$

The point-to-surface (P2S) distance serves as a metric to quantify the distance between a point and a surface within a 3D space. When employing the Euclidean distance, the P2S distance from a point $x$ to a surface $S$ can be formulated as $\min_{y \in S} \|x - y\|_2^2$, where $y$ denotes each point on $S$.

Given their definitions, the Chamfer distance and P2S distance are commonly utilized to evaluate 3D human reconstruction and human surface generation methods [117, 141, 142, 183, 184].

## 5.3 Qualitative and Quantitative Comparison of Representative Methods

Following the introduction of datasets and evaluation metrics, we present a qualitative and quantitative comparison of representative human image generation methods.

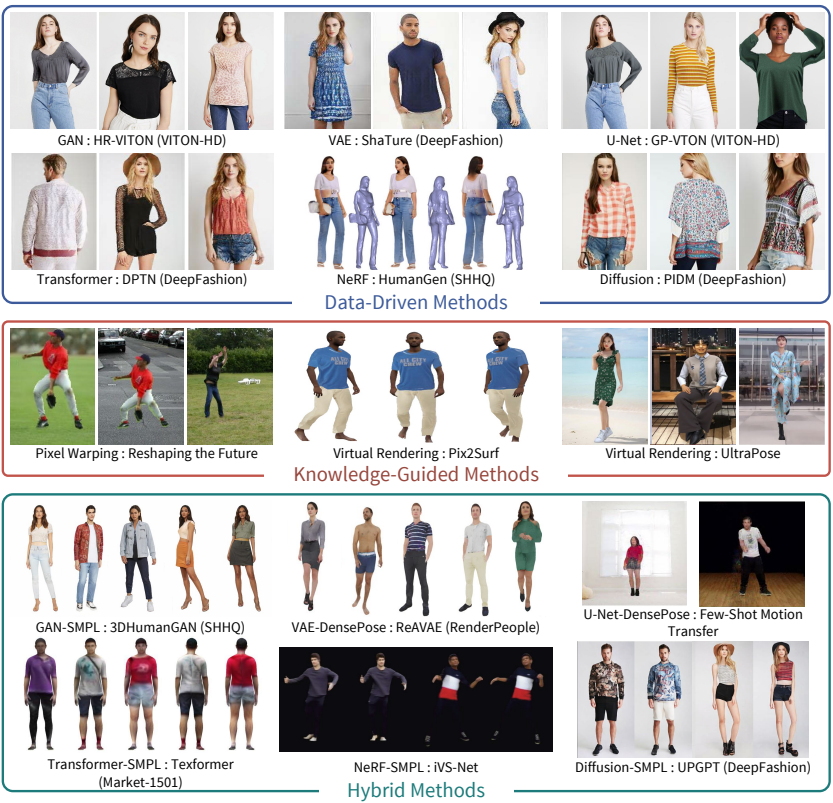

Fig. 12. The qualitative comparison of generated results of representative data-driven methods, knowledge-guided methods, and hybrid methods. The synthesized images are adapted from articles [15, 19, 27, 36, 65, 74, 83, 116, 129, 180, 190, 191, 201, 205, 213].

For the qualitative comparison, Figure 12 demonstrates generated human images from various method categories, each representing relatively recent work in its category. Method categories and names are labeled below the images, with datasets indicated in brackets. As the generated images of data-driven methods indicates, the GAN, U-Net, Transformer, and diffusion model-based methods are able to synthesize high-fidelity results. In contrast, images generated by NeRF-based methods may lack details. However, they excel in synthesizing 3D geometry of target person, enabling full-view synthesis. Additionally, the choice of datasets influences synthesis quality across methods. Regarding the knowledge-guided methods, these methods are able to synthesize human images with various poses and shapes based on the prior 3D human knowledge. However, artifacts may be noticeable in the generated images. Hybrid methods combine the strengths of data-driven and knowledge-guided methods, enabling synthesis of whole-body images with diverse poses, shapes, and photorealistic appearances. In summary, data-driven methods excel in learning dataset distributions, making them suitable for tasks requiring high-resolution and high-fidelity results [15, 28, 75, 83, 89, 180, 192, 212]. The hybrid methods leverage 3D human model for whole-body synthesis with pose and shape manipulations, making them ideal for human body video generation tasks [36, 37, 98, 204].

Pose-guided image generation and virtual try-on are among the most common tasks in human image generation research. In Table 4, we present a quantitative comparison of recent state-of-the-art methods, focusing on those proposed after 2021. The comparison utilizes evaluation metrics IS, SSIM, FID, and LPIPS, along with three representative datasets: Market-1501 [221], commonly used for evaluating pose-guided image generation; DeepFashion [104], which reflects the capabilities for both pose-guided image generation and virtual try-on tasks; and VITON-HD [28], primarily used for high-resolution virtual try-on. In the table, ↑ indicates higher values are better, while ↓ indicates the opposite. The table highlights those methods such as [15, 86, 87, 150] consistently achieve superior quantitative results in pose-guided image generation. Methods like [28, 89, 192] are better suited for high-resolution virtual try-on tasks.

Table 4. The Quantitative Comparison of Representative Human Image Generation Methods.

| Method Paradigms | Methods | Market-1501 | | | | DeepFashion | | | | VITON-HD | | | |
|---|---|---|---|---|---|---|---|---|---|---|---|---|---|
| | | IS ↑ | SSIM ↑ | FID ↓ | LPIPS ↓ | IS ↑ | SSIM ↑ | FID ↓ | LPIPS ↓ | IS ↑ | SSIM ↑ | FID ↓ | LPIPS ↓ |
| Data-Driven Methods | PoT-GAN [87] | **3.946** | **0.394** | 54.561 | - | 3.365 | 0.775 | 18.529 | - | - | - | - | - |
| | MUST-GAN [110] | - | - | - | - | **3.692** | 0.742 | 15.902 | 0.241 | - | - | - | - |
| | SCM-Net [175] | - | - | - | - | 3.632 | 0.751 | 12.18 | 0.182 | - | - | - | - |
| | DiOr [31] | - | - | - | - | - | 0.725 | 13.10 | 0.229 | - | - | - | - |
| | STTEN [187] | - | - | - | - | - | 0.774 | 9.888 | 0.182 | - | - | - | - |
| | APATN [230] | 3.426 | 0.326 | - | - | 3.184 | **0.784** | - | - | - | - | - | - |
| | DRN [198] | - | - | - | - | 3.125 | 0.774 | 14.611 | 0.218 | - | - | - | - |
| | ShaTure [205] | - | - | - | - | 3.487 | 0.777 | 11.434 | 0.159 | - | - | - | - |
| | SPGNet [106] | - | 0.315 | 23.331 | 0.278 | - | 0.782 | 12.243 | 0.211 | - | - | - | - |
| | CASD [226] | - | - | - | - | - | 0.725 | 11.373 | 0.194 | - | - | - | - |
| | PIDM [15] | - | 0.305 | 14.451 | 0.242 | - | 0.731 | **6.367** | 0.168 | - | - | - | - |
| | DPTN [213] | - | 0.285 | 18.995 | 0.271 | - | 0.778 | 11.466 | 0.196 | - | - | - | - |
| | WaveIPT [107] | - | - | - | - | - | 0.780 | 8.826 | 0.196 | - | - | - | - |
| | PoCoLD [57] | - | - | - | - | - | 0.731 | 8.067 | 0.164 | - | - | - | - |
| | PCDM [150] | - | 0.317 | **13.897** | 0.224 | - | 0.744 | 7.473 | **0.137** | - | - | - | - |
| | SAL-VTON [192] | - | - | - | - | - | - | - | - | - | **0.907** | 9.52 | 0.048 |
| | HR-VITON [83] | - | - | - | - | - | - | - | - | - | 0.892 | 10.91 | 0.065 |
| | VITON-HD [28] | - | - | - | - | - | - | - | - | - | 0.895 | 11.74 | **0.053** |
| | KGI [89] | - | - | - | - | - | - | - | - | - | 0.900 | **6.93** | 0.066 |
| Hybrid Methods | wFlow [37] | - | - | - | - | - | 0.844 | 57.652 | 0.187 | - | - | - | - |
| | Pose with Style [2] | - | - | - | - | - | 0.778 | 8.745 | **0.134** | - | - | - | - |
| | SPATT [100] | - | - | - | - | 3.39 | 0.779 | - | 0.161 | - | - | - | - |
| | HumanGAN [144] | - | - | - | - | - | 0.777 | - | 0.187 | - | - | - | - |
| | Texformer [190] | - | 0.742 | - | 0.115 | - | - | - | - | - | - | - | - |
| | PT$^2$ [86] | 2.789 | - | 17.389 | - | 3.469 | 0.795 | 8.338 | 0.158 | - | - | - | - |
| | UPGPT [27] | - | - | - | - | - | 0.697 | 9.427 | 0.189 | - | - | - | - |

An intriguing observation in Table 4 is the performance discrepancy in different metrics among certain methods, where they may excel in some metrics while perform inferiorly in others, e.g., PoT-GAN [87] and wFlow [37]. PoT-GAN [87] achieves the state-of-the-art IS and SSIM performance on Market-1501. However, its FID score is much higher than other methods. The reason is that images in Market-1501 contain complex backgrounds. Compared to the other methods [15, 106, 150, 213], PoT-GAN [87] is not proficient at synthesizing complex backgrounds, resulting in inferior FID performance. Regarding wFlow [37], it achieves a much higher FID score than other methods on DeepFashion. The reason is that they are reported for different generation tasks. wFlow [37] reports FID for garment-conditioned image transfer task, while the rest methods in Table 4 report FID for pose-conditioned image transfer task. Different experiment setups may also influence the quantitative performance. For example, on the DeepFashion dataset, some methods [31, 87, 106, 110, 187, 226] are trained and tested at a resolution of 256 × 256, while others [15, 57, 107, 150, 198, 205, 213] implement images at 256 × 176 resolution. The latter methods generally achieve superior FID and LPIPS performance. It's also worth noting that while the quantitative differences between some methods may be small, e.g., the SSIM scores on DeepFashion, these metrics alone may not comprehensively reflect the overall performance, model size, and computational cost of the methods. Researchers should choose comparable methods based on the specific requirements of the tasks and experiment setups.

## 6 APPLICATIONS

In the past four years, the metaverse has emerged as a focal point of artificial intelligence research, with human image and video generation playing a pivotal role in its development. As virtual humans show vast application potential in augmented reality, virtual reality, and the metaverse, numerous technology companies have unveiled projects focused on virtual humans or "MetaHuman", as summarized in Table 5.

Table 5. The Projects on MetaHuman of Technology Companies.

| Companies | Platforms | MetaHuman Characters | Companies | Platforms | MetaHuman Characters |
|---|---|---|---|---|---|
| **Nvidia** | Omniverse Avatar | Toy-Me | **Digital Domain** | Project Digi Doug | Digi Doug |
| **Meta** | Oculus Avatars | - | **Soul Machines** | - | YUMI |
| **Google** | The Relightables | - | **Baidu** | Xiling | XiaoC |
| **Microsoft** | Xiaoice Avatar Framework | Xiaoice | **DataGrid** | INAI MODEL | imma |
| **Samsung** | NEON | - | **SenseTime** | SenseMARS Agent | Xiaotang |
| **Epic Games** | MetaHuman Creator | - | **Genies** | Avatar Ecosystem | - |
| **Tencent** | xFaceBuilder & Matt AI | Siren | **Netease** | Youling MetaHuman | Lin Yaoyao |

Data augmentation is also an important application of the human image generation methods. Large-scale data is vital for deep learning models in human-related computer vision tasks. However, the data capture process for tasks like 3D human pose estimation is expensive and time-consuming due to the need for detailed ground truth pose annotations. To alleviate this challenge, researchers utilize human image generation models [22, 129, 138] to synthesize training data, enhancing the performance of these models. For pedestrian detection, Pedestrian-Synthesis-GAN [125] generates pedestrians with diverse appearance details on background images. Another application gaining popularity is in person Re-ID tasks. Various methods [9, 12, 94, 108, 159, 173, 222] have emerged to synthesize human images, enriching datasets and improving the accuracy and robustness of person Re-ID models. Furthermore, several person Re-ID datasets have been established with the help of knowledge-guided methods. The statistics information of these datasets is concluded in Table 6.

Table 6. The Statistics Information of Synthesized Person Re-ID Datasets.

| Datasets | Identities | Bounding Boxes | Cameras | Datasets | Identities | Bounding Boxes | Cameras |
|---|---|---|---|---|---|---|---|
| **SOMAnet** [12] | 50 | 100,000 | - | **RandPerson** [173] | 8,000 | 228,655 | 19 |
| **SyRI** [9] | 100 | 56,000 | - | **UnrealPerson** [215] | 3,000 | 120,000 | 34 |
| **PersonX** [159] | 1,266 | 273,456 | 6 | | | | |

Different from typical data augmentation applications, virtual try-on is deeply integrated into daily life, making it a highly sought-after research area. As depicted in Figure 3, virtual try-on tasks can be further categorized into garment image-conditioned VITON, person image-conditioned VITON, and pose-guided VITON. Given practical scenarios, real-time pose-guided VITON is particularly essential for applications like clothing e-commerce and customization. Numerous software solutions and applications, such as *triMirror*, *Cloudream*, *FXMirror*, *kitemiru*, and *Zeekit*, are actively pursuing the realization of this technology.

Upon analyzing the human image generation methods outlined in this survey, we discover that the segment-texture generation procedure aligns well with the garment-conditioned image transfer task. On the other hand, the warping-fusion approach finds suitability in the hybrid conditional image transfer task, particularly in scenarios like pose-guided VITON. As for the coarse-to-fine and texture-rendering generation procedures, they generally fit both data augmentation and virtual try-on tasks.

# 7 CHALLENGES AND OPPORTUNITIES

Despite the considerable attention given to human image generation, numerous challenges, as well as opportunities for further research persist in this domain.

- **Further improve the generation quality.** While significant progress has been made in human image generation methods, the qualitative generation effect still falls short of being entirely satisfactory, particularly in terms of photorealism. On one hand, a limited number of methods [28, 83, 89, 180, 192] have addressed the resolution challenge in human image generation. Many generated images remain in low resolutions due to several factors. Firstly, certain training datasets [34, 56, 104, 221] contain images of low resolution. Secondly, models designed for high-resolution image generation are usually large-scale and extremely entail computing resources. To address this, it is imperative to establish more large-scale, high-resolution datasets tailored to specific human image generation tasks. Additionally, prevalent super-resolution methods can be applied in both pre- and post-processing stages of human image generation. Various neural networks dedicated to general high-resolution image synthesis [41, 52, 139] could also play a pivotal role in enhancing human image generation quality. On the other hand, finer details such as facial features, hair styles, and garment textures are often lacking in synthesized images. To address this limitation, further exploration of methods within the disentanglement pipeline is required.
- **Pay more attention to human video generation.** After the advancements in still image generation, human video generation has emerged as a significant research area. Numerous methods [35, 37, 73, 75, 123, 194, 204, 207, 213] and datasets [25, 35, 37] have been introduced for human video generation. However, challenges remain in achieving high-resolution fidelity and real-time generation of human videos. Addressing these challenges will facilitate the practical applications of human image and video generation methods.
- **Propose more appropriative evaluation metrics.** As outlined in Section 5.2, manual rating stands out as the most appropriate method for evaluating image generation quality. However, it is impractical to manually review all generated human images across different methods. While manual rating offers qualitative insights, other commonly used quantitative evaluation metrics may not intuitively reflect the generation effect. Developing appropriate metrics is essential not only for the evaluation process but also for designing effective loss functions, thereby enhancing the generation capability of models. Therefore, proposing suitable and specific evaluation metrics for each task is crucial for advancing human image generation research.

- **Develop more interpretable human image generation.** The concept of interpretability in deep learning, or explainable artificial intelligence, has garnered significant attention in the research community. However, most human image generation methods treat generative networks as black boxes. Several studies have attempted to synthesize human images in a disentangled manner, enabling control over various dimensions of generated attributes [10, 47, 101, 109, 113, 121, 144]. To achieve a fully controllable generation process, further research on the interpretability of human image generation is essential.
- **Promote the efficiency of human image generation models.** Improving the efficiency of models is a persistent challenge across various computer vision and computer graphics tasks, including human image generation. Model compression and acceleration are vital for the practical deployment of deep learning models. In particular, virtual try-on applications, especially pose-guided VITON, hold significant value and are likely to be extensively used in various e-commerce scenarios. Real-time models with high efficiency are crucial for these applications. However, there has been limited focus on developing virtual try-on models suitable for real-time deployment. Models like WUTON [69] and PF-AFN [48] address this challenge by employing semantic parsers with knowledge distillation frameworks, ensuring efficiency during model inference. More efforts are needed to further enhance the efficiency of virtual try-on models.
- **Provide privacy-free human image data.** Addressing the privacy concerns surrounding data is a pressing issue within the field of artificial intelligence, particularly in human-related tasks. Human image generation presents a potential solution to this problem. In the context of person Re-ID, several synthesized datasets have been introduced, as outlined in Table 6. There is a hope that similar efforts can be extended to other human-related tasks, such as human attribute detection and person detection and retrieval.

## ACKNOWLEDGMENTS

This work is jointly supported by the National Science and Technology Major Project (Grant No. 2022ZD0117901), the National Natural Science Foundation of China (Grant No. 62106260, 62373355 and 62236010), and the China Postdoctoral Science Foundation (Grant No. 2020M680751).

Human Image Generation: A Comprehensive Survey 1:35